\documentclass[11pt,a4paper]{article}
\usepackage[hyperref]{acl2019}
\usepackage{times}
\usepackage{latexsym}

\usepackage{times}  
\usepackage{helvet}  
\usepackage{courier}  
\usepackage{url}  
\usepackage{graphicx}  
\usepackage{amsmath,amsfonts,amssymb}
\usepackage{subcaption}
\usepackage{multirow}
\usepackage{pgfplots}
\usepackage[symbol]{footmisc}
\graphicspath{{./}}

\usepackage{url}

\aclfinalcopy 

\begin{document}

\author{Sajad Movahedi\thanks{\hspace{0.2cm} Equal Contribution.}, Erfan Ghadery\footnotemark[1], Heshaam Faili, Azadeh Shakery\\
University of Tehran\\
Tehran, Iran\\
\{s.movahedi, erfan.ghadery, hfaili, shakery\}@ut.ac.ir\\
}

\date{}

\title{Aspect Category Detection via Topic-Attention Network}

\maketitle

\begin{abstract}
E-commerce has started a new trend in natural language processing through sentiment analysis of user-generated reviews. Different consumers have different concerns about various aspects of a specific product or service. Aspect category detection (ACD), as a subtask of aspect-based sentiment analysis, tackles the problem of categorizing a given review sentence into a set of pre-defined aspect categories. By nature, in this task, a given review sentence can belong to one or more categories.
In recent years, the attention mechanism has brought revolutionary advances in multiple branches of natural language processing including sentiment analysis by attending to informative words or phrases in the text. However,  in multi-label classification tasks, such as ACD, given different labels, we need to attend on different parts of a given sentence, which is not addressed by the vanilla attention methods. 
In this paper, we propose a deep neural network method based on attention mechanism to identify different aspect categories of a given review sentence by attending to various parts of the review sentence based on different topics, which are more fine-grained than aspects categories.  Experimental results on two datasets in the restaurant domain released by SemEval workshop demonstrates that our approach outperforms existing methods on both datasets. Visualization of the topic attention weights shows the effectiveness of our model in identifying words related to different topics.
\end{abstract}

\section{Introduction}
User-generated reviews in e-commerce websites are valuable resources for both the consumers and the producers of products or services. For a potential consumer, the experiences of other consumers help making educated decisions before purchasing a product or service. On the other hand, such data can help the producers of these products or services in refining the quality of what they offer. Different customers can have different concerns about the same product. This issue raises the challenge of categorizing the reviews into pre-defined aspects of the product under review. This challenge is tackled by ACD, a subtask of aspect-based sentiment analysis. Given a review sentence, ACD aims to categorize the sentence into a set of pre-defined categories like `FOOD', `PRICE' etc.  in the restaurant review domain. For example, the sentence ``It is very overpriced and not very tasty." belongs to both `FOOD' and `PRICE' aspect categories.

Due to the multi-label nature of this task, most of the approaches utilize one-vs-all classification models. These methods have shown good results in performing aspect category detection. However, training several one-vs-all classifiers require a lot of resource and time, especially when there are numerous categories. Another issue is raised by the fact that for different aspect categories, different words may contribute variously. Attention mechanism \cite{bahdanau2014neural} has shown promising performance in aspect-based sentiment analysis \cite{he2017unsupervised} by attending to informative words or phrases in the text. However, in multi-label classification tasks, such as ACD, we need to attend to different parts of a sentence when considering different aspect categories. Nevertheless, each aspect category is a combination of multiple topics; for example, ``ambiance'' aspect category in the restaurant domain is a combination of topics such as ``temperature'', `` beauty'', etc. Therefore, we can detect aspects by recognizing their constituent topics. In this paper, instead of training several one-vs-all models, we propose a single model, namely Topic-Attention Network (TAN), which can detect aspect categories of a given review sentence by attending to different parts of the sentence based on different topics.

\par The common practice in the literature for text classification tasks is to maximize a scalar value as the probability of the corresponding class in the optimization process. Inspired by the \cite{sabour2017dynamic}, instead of maximizing a scalar value, we propose to learn a vector for each aspect category and maximize the length of these vectors. Therefore, we borrow the squash activation function proposed by \cite{sabour2017dynamic} to get a vector with a normalized length of at most 1, since we treat the lengths as probabilities. We hypothesized that breaking a vector to a scalar value will lose some of the orientational and length related information contained by the vector. In the experiments, we show that treating the problem in the aforementioned way will cause an improvement in the performance of the model. 

\par Inspired by \cite{he2017unsupervised}, we utilize a regularization term to preserve the orthogonality of the weights corresponding to each topic. We believe this regularization term will help the diversity of the topics, allowing the model to become more efficient in its use of the topics.

\par Using a bi-directional GRU layer, TAN obtains encoded representations of each word. These representations are then fed to the topic attention layer to acquire attentive representations of the sentence. Attentive sentence representations each are fed to a fully connected layer. Following the same hypothesis mentioned above, we utilize the squash function for each of the attentive representation in order to preserve the length related and orientational information of each topic. Then, the squashed vectors obtained from the previous layer are concatenated together to provide a multi-topic representation of the sentence. For each aspect category, we feed the sentence representation into a fully connected layer followed by squash. The length of each vector is treated as the probability of its corresponding aspect. If the probability of an aspect category surpasses a threshold, the aspect category will be assigned to the review sentence.

\par We evaluate our proposed method by comparing it with several baselines in two freely available benchmark datasets of SemEval workshops. The results confirm the effectiveness of the proposed method, and visualization of the topic-attention weights shows that TAN is able to efficiently attend to different parts of a sentence, given different topics.

\par Our contributions in this paper are as follows. 
\begin{itemize}
    \item We propose to attend to the review sentences based on topics, which is more fine-grained compare to aspects
    \item In order to preserve the orientational and length related information, we formulate the problem as learn a vector for each aspect category and maximize the length of the vectors corresponding to each aspect, instead of maximizing scalar values
    \item We utilize a regularization term inspired by \cite{he2017unsupervised} to preserve the orthogonality of the weights corresponding to each topic in the attention mechanism
\end{itemize}

\section{Related Work}
\par Previous research works in this field can be divided into five approaches: frequency-based, syntax-based, supervised machine learning, unsupervised machine learning, and hybrid \cite{schouten2016survey}. The majority of the proposed approaches are machine learning based including classic algorithms such as SVM and Maximum Entropy \cite{xenos2016aueb}, \cite{hercig2016uwb}, and deep neural network based approaches \cite{toh2016nlangp}, \cite{xue2017mtna}.
Aspect-based sentiment analysis has gained much attention in recent years following the pioneering work of \cite{hu2004mining}. Based on a hypothesis that aspects are nouns or noun phrases, they used an association rule mining to extract frequent nouns and noun phrases as the candidates for aspects. In the next step, a set of rules are conducted to prune non-aspect candidates. \cite{qiu2011opinion} proposed to use a double propagation technique to extract aspect terms and opinion terms in an iterative manner. They conduct a set of rules based on dependency relations to extract aspect terms from opinion terms like `good' or `bad', and vice versa.
\par Aspect category detection is a subtask of aspect-based sentiment analysis, which instead of extracting aspect terms, there are a set of pre-defined aspect categories like `FOOD' and `PRICE', and the goal is assigning a subset of these categories to a given review sentence. SemEval workshop has addressed aspect category detection subtask for three consecutive years, which attracted a lot of contestants, especially in SemEval 2016 \cite{pontiki2016semeval}. \cite{kiritchenko2014nrc} proposed multiple features including n-grams, lexicon features, etc. to train a set of one-vs-all SVM classifiers. This model was the top contestant of SemEval 2014 \cite{semeval2014}. In \cite{xenos2016aueb}, authors train a set of one-vs-all SVM classifier with several hand-crafted features. By calculating the Precision, Recall, and F1-Score of the stemmed and un-stemmed N-grams on the train data, they create a set of lexicons for providing features to the classifiers. 
\par In recent years, deep neural network based approaches have been used to address the aspect category detection task, achieving state-of-the-art results. In \cite{toh2016nlangp}, authors proposed using the output of a convolutional neural network trained on the dataset as features for a set one-vs-all of linear classifiers along with several other features such as ngrams and POS tags. This work was the top contestant in SemEval 2016. In \cite{zhou2015representation} two other loss functions were added to the skip-gram model introduced by \cite{mikolov2013distributed} to train a word embedding specifically for aspect category detection. Using a set of multi-layer perceptrons, a set of hybrid features were extracted on the average of word embeddings to train another set of one-vs-all classifiers to extract aspect categories. In \cite{xue2017mtna} a set of one-vs-all deep neural networks composed of a CNN layer on top of an LSTM layer was proposed to be trained on both aspect category and aspect term labels simultaneously. A deep neural network approach based on attention mechanism is proposed in \cite{he2017Attention}. In this paper, the authors use a network similar to an autoencoder in order to perform unsupervised aspect category detection. The proposed network is trained in a way that attends to the aspect-relevant terms.

\par The most similar work to ours is the one proposed in \cite{hu2018can}. In this paper, the authors propose a regularization term to encourage the orthogonality of the weights corresponding to the attention heads. Compared to our method, the sentence representation proposed in this paper is based on aspects, and the probability of the sentence belonging to each aspect is only related to its corresponding aspect-related representation. Furthermore, the model proposed in this paper performs aspect category detection and aspect jointly trained on the aspect category and aspect sentiment classification jointly, while our focus is on aspect category detection.

\section{Topic-Attention Network}
Figure 1 represents the architecture of Topic-Attention Network (TAN). Our network consists of several components including a sentence encoder layer, a topic-attention layer, and two non-linear transformation layers. In the following subsections, we describe the details of different parts of our model. 
\begin{center}
\begin{figure}[t]
    \centering
    \includegraphics[width=0.5\textwidth]{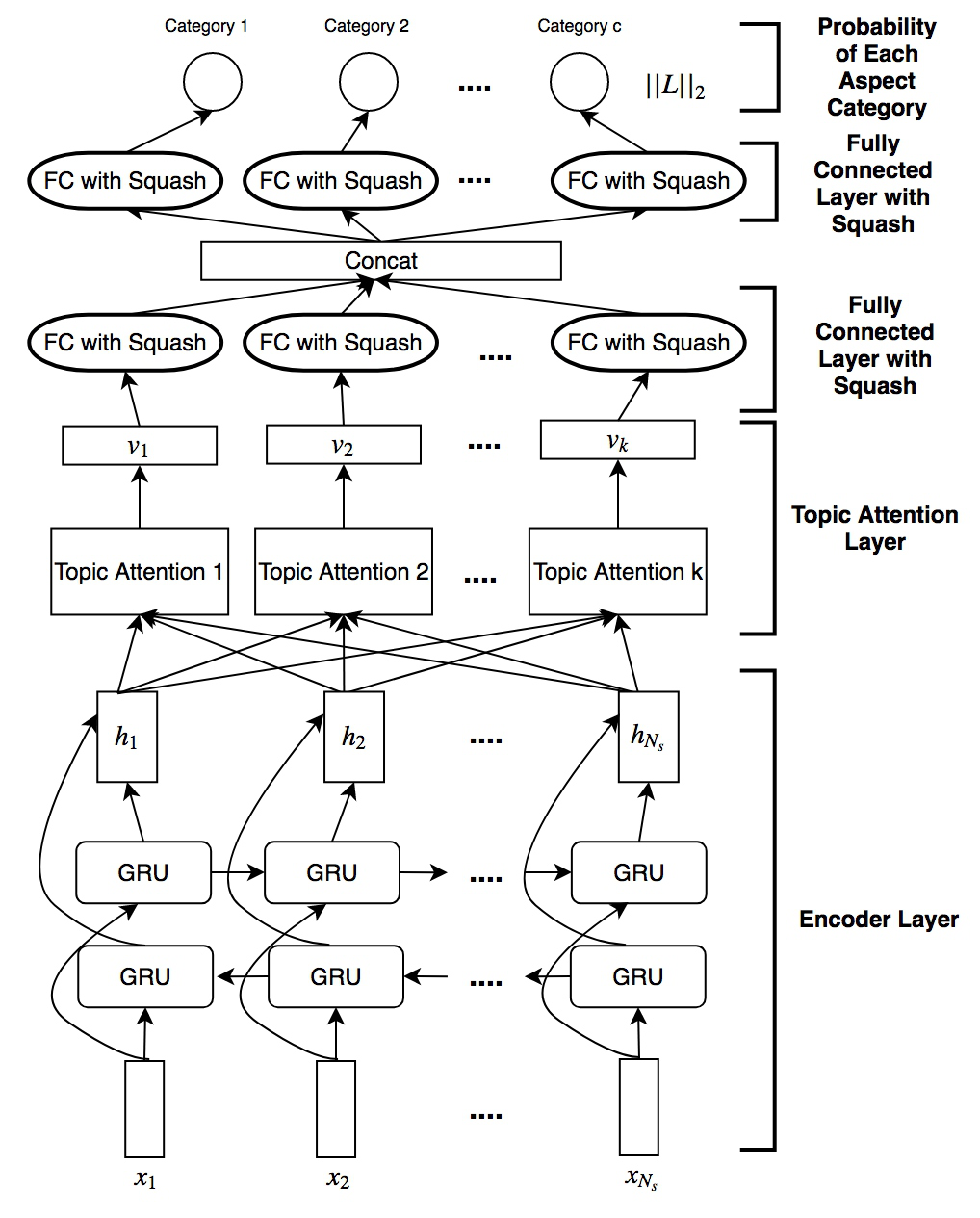}
    \caption{The architecture of TAN.}
\end{figure}
\end{center}
\subsection{Sentence Encoder Layer}
We employed a bi-directional recurrent neural network to extract the sequential information for each word. We utilized the Gated Recurrent Unit (GRU) \cite{cho2014learning} for this purpose. In \cite{chung2014empirical} the GRU was found to show better performance on small datasets.  In order to track the state of sequences, GRU utilizes a gating mechanism without using separate memory cells. The GRU is formulated as follows:
\begin{equation}
     r_t = \sigma({W_i}_r x_t + {b_i}_r + {W_h}_r h_{(t-1)} + {b_h}_r) 
\end{equation}
\begin{equation} 
    z_t = \sigma({W_i}_z x_t + {b_i}_z + {W_h}_z h_{(t-1)} + {b_h}_z) 
\end{equation}
\begin{equation} 
    n_t = \tanh({W_i}_z x_t + {b_i}_z + {W_h}_z h_{(t-1)} + {b_h}_z) 
\end{equation}
\begin{equation} 
    h_t = (1-z_t)\odot n_t + z_t \odot h_{(t-1)}
\end{equation}
where $x_t$ is the input, $r_t$ is the reset gate, $z_t$ is the update gate, $n_t$ is the candidate state, and $\odot$ is pair-wise multiplication. The reset gate $r_t$ decides how of the past state $h_{(t-1)}$ is used in the current state $h_t$ and the update gate $z_t$ decides how much of the candidate state $n_t$ is used in the current state $h_t$.
\par The input of the encoder layer is a continuous representation of the input sentence using word embeddings.

\subsection{Topic-Attention Layer}

Not all the words in a sentence contribute equally to the meaning of the sentence. Furthermore, when a review sentence is expressing opinions about different aspects of a product, this variation in the importance of the words become relevant to the aspects. For example, consider the review sentence ``It is very overpriced and not very tasty." in the restaurant domain. In this example, the word `overpriced' is more important than the other words if the given aspect category is `price'. On the other hand, if the given aspect category is `food', the word `tasty' should be given higher importance compared to the other words. 

The attention mechanism proposed in \cite{bahdanau2014neural} aims to solve this exact problem. The attention mechanism proposed in \cite{bahdanau2014neural} helps the model to attend to different parts of the source sentence based on the previously detected token of the target sentence. Inspired by this idea, we propose to attend to a given review sentence based on different topics. 

Because of the multi-label nature of this task, a single attention may not be able to provide a good representation of the sentence in cases where the review sentence contains multiple aspects. Furthermore, an aspect may be too general and represent multiple characteristics of the product under review. For example, the aspect category `FOOD' references to many products served in a restaurant, from different meals to salads, desserts, and even different types of sauce. Inspired by the target-based attention proposed in \cite{bahdanau2014neural}, we propose to solve this problem by attending to the review sentence based on several topics, each refering to one or more aspects. 

\par Given the $i^{th}$ topic and word representations$\{h_{t}\}_{ t= 1}^{N_{s}}$ obtained from the encoder layer, the final attentive sentence representation $v_i$ is computed as follows:
\begin{equation}
    {e_i}_t = h_t T_i
\end{equation}
\begin{equation}
    {\alpha_i}_t = \frac{exp({e_i}_t)}{\sum_{j=1}^{N_s} exp({e_i}_j)}
\end{equation}
\begin{equation}
    v_i = \sum_{j=1}^{N_s} {\alpha_i}_j h_j
\end{equation}
where ${e_i}_t$ determines the importance of the $t^{th}$ word with respect to $i^{th}$ topic by measuring the similarity between the word and the topic weight $T_i$, which is trained as part of the weights of the model, and ${\alpha_i}_t$ is the normalized value of ${e_i}_t$, using the softmax function, where $N_s$ is the length of the given sentence. Finally, we compute attentive sentence representation $v_i$ as a weighted sum of word representations obtained from the encoder layer based on their attention weights. Intuitively, the attention weight $T_i$ plays the role of a filter, which learns to emphasize the relevant words to the topic. The number of topics is one of the hyperparameters of the network and the weights corresponding to each topic is trained during the training process.

\subsection{Non-Linear Transformation with Squash}
We utilize the squash function proposed in \cite{sabour2017dynamic} which is a non-linear function that ensures the length of almost zero for short vectors and a length of slightly below 1 for long vectors. Given an input $x \in R^m$ where $m$ is the length of the input vector, the output of the squash function for $x$ can be formalized as follows:
\begin{equation}
    squash(x) = \frac{||x||^2}{1 + ||x||^2} \frac{x}{||x||}
\end{equation}
\par In this layer, each output vector obtained from the topic attention layer is fetched into a one-layer MLP in order to extract high-level features and also perform dimensionality reduction simultaneously. The non-linear squash function is then applied to the output vector of the MLP in order to reduce the length of the vector to be no more than 1 while preserving the direction of the vector.
\par All k squashed vectors are then concatenated together to provide features for the next similar layer.  
\subsection{Category Detection}
The output of the fully connected with squash layer is a squashed vector with the length between 0 and 1, which its length is considered as a probability value. Therefore, the last layer of the TAN consists of c parallel fully connected layers with squash where c is the number of categories. To model the probability that a review sentence belongs to each of the categories, we calculate the L2-Norm of the output vector corresponding to each category.  In other words, the longer the length of an output vector, the higher the probability of the given sentence belonging to the corresponding category.
\subsection{Training Objective}
The training parameters of our model ($\theta$) consists of the weights of the model and the weights of the topics.
\par We utilize the Mean Square Error loss formalized as follows:
\begin{equation}
    \mathcal{J}(\theta) = \frac{1}{n} \sum_{i=1}^n (y_i - \hat{y_i})^2
\end{equation}
where $n$ is the number of elements that the loss function is applied to, $y_i$ is the ground truth value of the $i$th element, and $\hat{y_i}$ is the predicted value for the $i$th element.
\par Following \cite{he2017unsupervised}, we also applied a regularization term to the model to keep the topic weights orthogonal and encourage the uniqueness of the topics. Given $T \in R^{k \times d}$ where $k$ is the number of topics and $d$ is the size of the topic weights, we define the regularization term $\mathcal{U}$ as follows:
\begin{equation}
    \mathcal{U}(\theta) = || T_{n} \cdot {T_{n}}^\intercal - I||
\end{equation}
where $T_n$ is the topic weights normalized to have a length of 1 and $I$ is the identity matrix. This term encourages all the non-diagonal elements of $T_n \cdot {T_{n}}^\intercal$ to have the value of zero. This means the dot product of the topic weights are encouraged towards being zero. Finally, our overall loss function will be:
\begin{equation}
    \mathcal{L}(\theta) = \mathcal{J}(\theta) + \mathcal{U}(\theta)
\end{equation}
\section{Experiments}
\subsection{Datasets}
For our experiments, we consider two datasets from SemEval workshop, SemEval-2014 task 4 \cite{semeval2014} and SemEval-2016 task 5 \cite{pontiki2016semeval}. SemEval 2015 dataset was not used because according to \cite{pontiki2016semeval}, it exists in SemEval 2016 dataset. In both datasets, we used restaurant domain reviews for our experiments. Table 1 shows the statistics of datasets. In SemEval-2016 there are 12 categories which are a combination of aspect and attribute pairs, (e.g. `Food\#Quality', `Service\#General'), while SemEval-2014 has 5 categories which are aspects only, (e.g. `Food', `Price'). Review sentences that don't contain any sentiments and therefore don't belong to any of the aspect categories are discarded in the training and validation data.
\begin{table}
    \begin{center}
    \begin{tabular}{c|c|c|c}
    Dataset & Train & Test & Total  \\ \hline
    SemEval-2014 & 3041 & 800 & 3841 \\ \hline
    SemEval-2016 & 2000 & 676 & 2676
    \end{tabular}
    \caption{The data statistics of the two datasets used for experiments. The numbers denote the number of sentences in each dataset.}
    \end{center}
\end{table}

\subsection{Baseline Methods}
In order to show the merit of our model, we compare TAN with multiple baselines in each dataset. The baseline methods are as follows:
\begin{itemize}
    \item \textbf{NLANGP.} The model introduced in \cite{toh2016nlangp} consists of a CNN model trained on word embeddings and a set of features including POS tags, word clusters, name lists which are fed to a set of binary classifiers for each category. This method is the top-ranked contestant in SemEval 2016 Aspect Category Detection subtask.
    \item \textbf{MTNA.} This method which was introduced in \cite{xue2017mtna} utilizes both aspect category and aspect term information to train a set of one-vs-all deep neural network models consisting of an LSTM layer followed by a CNN layer.
    \item \textbf{NRC-Canada.} The model proposed in \cite{kiritchenko2014nrc} is the top-ranked contestant in SemEval 2014 aspect category detection subtask. This model addresses the aspect category detection task using a set one-vs-all SVM classifiers (one classifier for each category) using several features including lexicon features, n-grams, word clusters, etc.
    \item \textbf{RepLearn.} In \cite{zhou2015representation} a semi-supervised in-domain approach for training word embeddings is introduced to capture the semantic relations between words, aspects, and sentiment words and aspects. Afterwards, multiple classifiers are trained to capture hybrid features, which are then used as features for a set of one-vs-all logistic regression classifiers to determine the aspect categories.
    \item \textbf{CAN.} This is the model proposed in \cite{hu2018can}. This model utilizes an LSTM layer to encode the input, and an attention mechanism for each of the aspects and sentiments to perform ACD jointly with aspect sentiment classification. They also use a regularization term similar to ours to preserve the orthogonality of the attention weights corresponding to each aspect and sentiment. They only performed ACD experiments SemEval 2014 dataset.
    \item \textbf{Vanilla-Attention(VA).} In order to demonstrate the effectiveness of utilizing multiple attentions in the network, we compare our method with a model consisting of an encoder layer, an attention layer, and a fully connected hidden layer with the ReLU activation function. The output of the hidden layer is then fed to another fully connected layer with the Sigmoid activation function to represent the probability of each aspect.
    \item \textbf{Topic-Attention-without-Squash(TAwS).} This baseline was added to demonstrate the effectiveness of treating the problem as learning vectors for aspect categories  and maximize the length of them instead of maximizing a scalar value for each category. Also, this baseline demonstrates
    the effectiveness of using non-linear squash activation function. In this model, the output of the topic attention layers in TAN are fed to a fully connected layer followed by the ReLU activation function and then concatenated. Then, the concatenated vector is fed to another fully connected layer with the Sigmoid activation function to represent the probability of each aspect.
\end{itemize}
\subsection{Experiment Settings}
In our experiments, we use F1-score, Precision, and Recall as evaluation measures. Stop-words and punctuation removal is performed as a preprocessing step using the NLTK package \cite{nltk}. For the input of TAN, TAwS, and VA, we train the word embeddings on the large unlabeled Yelp challenge dataset \footnote{https://www.yelp.com/dataset/challenge} using the genism package \cite{gensim} implementation of the skip-gram model introduced in \cite{mikolov2013distributed}. We set the size of the word embeddings to 300 and the rest of the parameters were set to default.

\par We also select 10 percent of the training data as the validation set for each aspect category in a uniform manner. All the hyperparameters of the model are tuned on the validation set using grid search. The optimum hidden size of the GRU is found to be 128 for which the combination of forward and backward GRU leads to a 256 dimension vector for each word annotation. The size of the weights of topics is therefore set to 256, and the optimum number of the topics was found to be 11 for SemEval-2016 dataset and 6 for SemEval-2014 dataset. Since SemEval-2014 has a relatively simpler data compared to SemEval-2016, this difference in their optimum topic number makes sense. We set the size of the squashed vectors of the first fully connected layer with squash to 32 and 64 for the next similar layer for SemEval-2016 dataset, and 16 and 32 for SemEval-2014 dataset.
\par For training the model, we use a mini-batch size of 128, and training is performed using the Adam optimizer \cite{adam}. We use drop out with the probabiliy of 0.6 in order to prevent the overfitting of our model. The model is trained for a maximum of 300 epochs for which early-stopping is performed with the patience set to 20.
\par We implemented VA, TAwS, and TAN using PyTorch \cite{paszke2017automatic} version 0.4.1. All the experiments were done on a GeForce GTX 1080.

\subsection{Results and Analysis}
The comparison results are shown in table 2. We extracted the results reported for MTNA, NRC-Canada, RepLearn, and NLANG from the original papers \cite{xue2017mtna}, \cite{kiritchenko2014nrc}, \cite{zhou2015representation}, and \cite{toh2016nlangp} respectively.
\par On the SemEval 2016 dataset, which has more fine-grained aspect categories and at the same time, is smaller compared to the SemEval 2014 dataset, our model comfortably surpasses the other baselines in terms of F1 score. Compared to the MTNA baseline which utilizes aspect term information in the training process, our method outperforms MTNA by 1.96\%. Compared to the other baselines, TAN also performs better, surpassing VA, TAwS, and NLANGP in term of F1 measure by 3.1\% and 3.54\%, and 5.3\% respectively. Interestingly,  VA baseline which is a vanilla version of TAN - TAN without multiple attention as topics - achieves better results compared to all the other baselines in the SemEval-2014 dataset, and outperforms TAwS and NLANGP in the SemEval-2016 dataset, which indicates the strength of the attention mechanism for aspect category detection. Also, we see that TAwS in both datasets achieves a worse result compared to VA, which confirms the effectiveness of utilizing squash function in our method. On the SemEval 2014 dataset, TAN also performs better than the other baselines in term of F1 measure. Our method outperforms VA, RepLearn, TAwS, MTNA, and NRC-Canada baseline by 0.44\%, 0.51\%, 0.81\%, 1.7\%, and 2.03\%, respectively. As can be seen in Table 2, CAN performs worse than even the top contender in SemEval 2014. We believe this poor performance can partly be attributed to CAN's use of word embeddings trained on out of domain data. But in any case, the performance of CAN using in-domain word embeddings probably would not be very different from the TAwS, given that similar structure is used in both of the models, and the same regularization term is utilized. Therefore, our model probably still would achieve better performance compared to CAN trained on in-domain word embeddings.

\begin{table}
\small
    \begin{center}
    \begin{tabular}{c|c|c|c|c}
    Dataset & Method & P (\%) & R (\%) & F1 (\%)  \\ \hline
    \multirow{7}{*}{SemEval 2014} 
    & CAN & 89.07 & 86.27 & 87.65 \\ \cline{2-5}
    & NRC-Canada & 91.04 & 86.24 & 88.58 \\ \cline{2-5}
    & MTNA & - & - & 88.91 \\ \cline{2-5}
    & TAwS & \textbf{93.24} & 86.41 & 89.70 \\
    \cline{2-5}
    & RepLearn & - & - & 90.10 \\ \cline{2-5}
    & VA & 91.54 & 88.85 & 90.17 \\ \cline{2-5}
    & \textbf{TAN} & 91.60 & \textbf{89.63} & \textbf{90.61} \\
    \hline
    \multirow{5}{*}{SemEval 2016}
    & NLANGP & 72.45 & 73.62 & 73.03 \\ \cline{2-5}
    & TAwS & 71.11 & 78.97 & 74.84 \\ \cline{2-5}
    & VA & \textbf{76.06} & 74.52 & 75.28 \\ \cline{2-5}
    & MTNA & - & - & 76.42 \\ \cline{2-5}
    & \textbf{TAN} & 74.78 & \textbf{82.34} & \textbf{78.38} \\ \hline
    \end{tabular}
    \caption{The experimental results of our method (TAN) compared with baselines.}

    \end{center}
\end{table}

\subsection{Visualization of Topic-Attention}
In this section, we visualize attention weights of sentence words for different topics. Note that, for each word we have a set of attention scores where each score shows the probability of the word belonging to a specific topic. Figure 3 shows the example of attention scores visualization for two sentences from both of the datasets. Each Column denotes a specific topic, so the sum of attention scores in every column is 1.
From Figure 3a we can see that there are several topics in the sentence. Topic 1 give an attention score of 1.0 to the word `service' and 0 to the other words, which shows that maybe this topic models the category `SERVICE\#GENERAL'. Similarly, the words `decor', `food', `delicious', `large', and `portions' get high attention scores each for a specific topic. Intuitively, these words represent different categories, so we expect the model to categorize these words in different topics. 
Figure 3b shows an example of SemEval-2014 dataset. This dataset is relatively simpler than the SemEval-2016 dataset so here we have 6 topics. According to the attention scores, we can see that obviously topics 1 and 2 models the drinks and foods respectively, which corresponds to the `FOOD' category, and topic 1 models the subjects related to the `Anecdote/Miscellaneous' category.      
\begin{center}
\begin{figure}[!t]
    \centering
    \begin{subfigure}[b]{0.5\textwidth}
    \includegraphics[width=\textwidth]{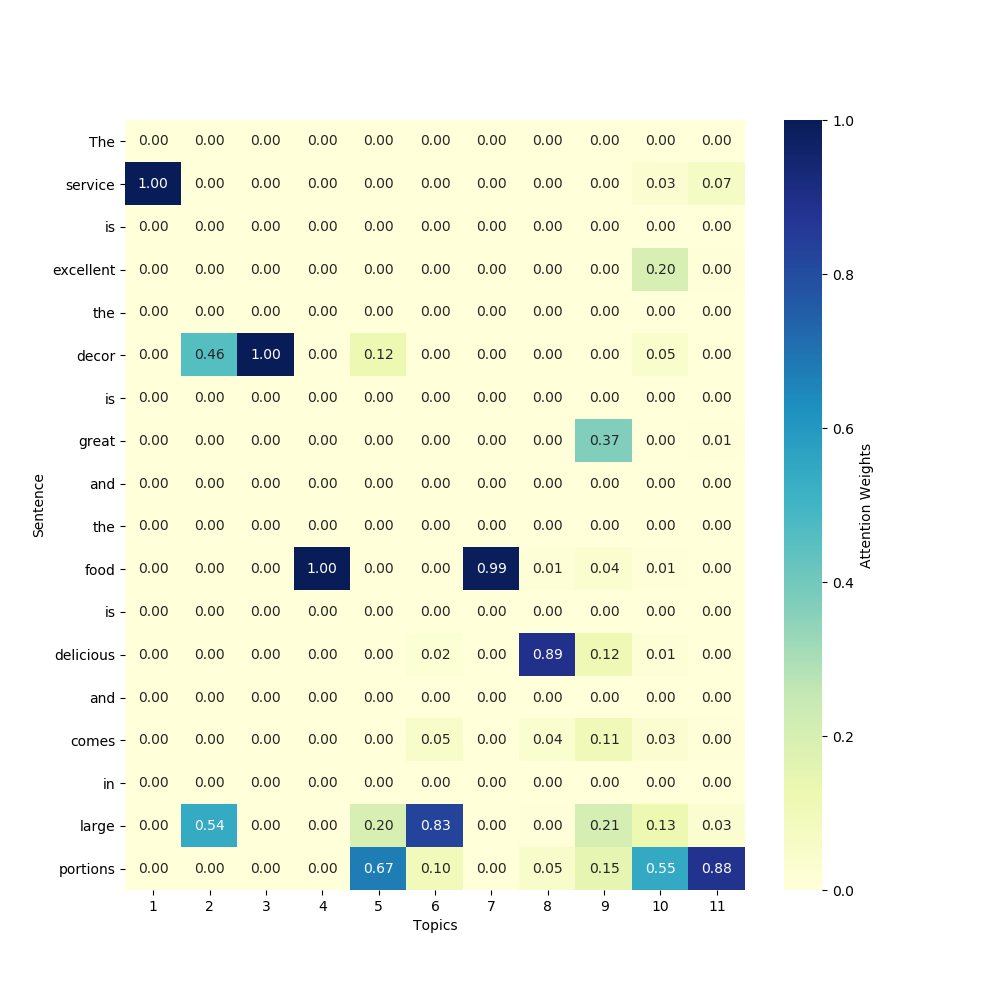}
     \caption{An example from SemEval 2016 restaurant dataset.}
    \end{subfigure}
    \begin{subfigure}[b]{0.5\textwidth}
    \includegraphics[width=\textwidth]{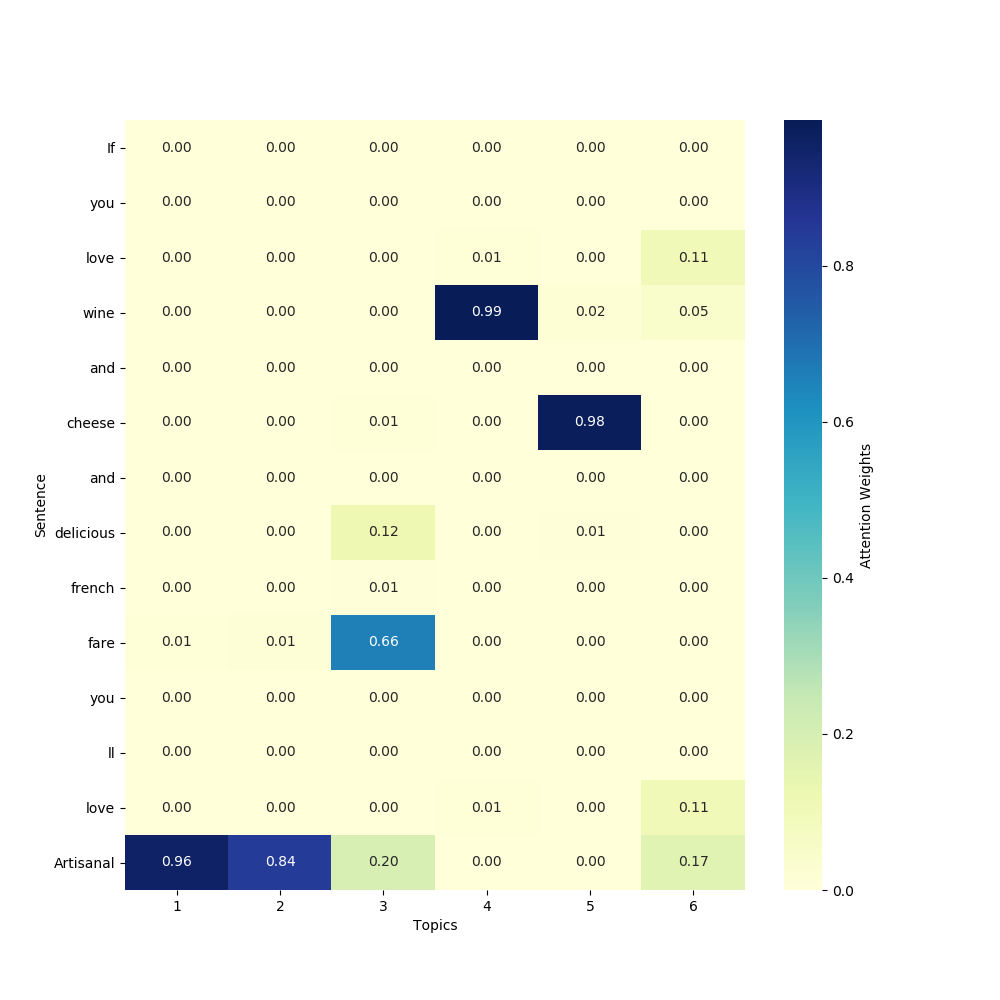}
     \caption{An example from SemEval 2014 restaurant dataset.}
    \end{subfigure}
    \caption{The visualization of the attention values.}
\end{figure}
\end{center}
\section{Conclusion}
In this paper, we propose a deep neural network based model composed of an encoder layer utilizing GRU recurrent units, a topic-attention layer producing sentence representations based on the existing topics in the data, and two fully connected layers that transfer representations into a vector space, using the squash activation function. Empirical results prove the effectiveness of our model compared to several baselines, including a single attention model and another version of our model without the squash activation function. This indicates the effectiveness of utlizing a topic attention mechanism and non-linear transformation in the vector space via the squash activation function.
\bibliography{acl2019}
\bibliographystyle{acl_natbib}

\end{document}